\documentclass[sigconf]{acmart}

\AtBeginDocument{%
  \providecommand\BibTeX{{%
    \normalfont B\kern-0.5em{\scshape i\kern-0.25em b}\kern-0.8em\TeX}}}

\setcopyright{acmcopyright}
\copyrightyear{2018}
\acmYear{2018}
\acmDOI{10.1145/1122445.1122456}

\acmConference[Woodstock '18]{Woodstock '18: ACM Symposium on Neural
  Gaze Detection}{June 03--05, 2018}{Woodstock, NY}
\acmBooktitle{Woodstock '18: ACM Symposium on Neural Gaze Detection,
  June 03--05, 2018, Woodstock, NY}
\acmPrice{15.00}
\acmISBN{978-1-4503-XXXX-X/18/06}
\usepackage{algorithmic}
\usepackage{balance}

\usepackage{booktabs}
\usepackage{multirow}
\usepackage{algorithmic}
\usepackage{enumitem}
\usepackage{makecell}
\usepackage{graphicx}
\usepackage{caption}
\usepackage{subfigure}
\usepackage{textcomp}
\usepackage{listings}
\usepackage{xcolor}

\usepackage{booktabs}
\usepackage{multirow}
\usepackage{algorithmic}
\usepackage{enumitem}
\usepackage{makecell}
\usepackage[linesnumbered, ruled, vlined]{algorithm2e}
\begin{document}

\title{ TCL: Transformer-based Dynamic Graph Modelling via Contrastive Learning
}


\author{Lu Wang}
\affiliation{\institution{East China Normal University}
\country{China}}
\email{luwang@stu.ecnu.edu.cn}

\author{Xiaofu Chang}
\affiliation{\institution{Damo Academy, Alibaba Group}\country{China}}
\email{changxiaofu123@163.com}

\author{Shuang Li}
\affiliation{\institution{Harvard University}\country{USA}}
\email{shuangli@fas.harvard.edu}

\author{Yunfei Chu}
\affiliation{\institution{Damo Academy, Alibaba Group}\country{China}}
\email{fay.cyf@alibaba-inc.com}

\author{Hui Li}
\affiliation{\institution{Ant Group}\country{China}}
\email{ken.lh@antgroup.com}

\author{Wei Zhang}
\affiliation{\institution{East China Normal University}
\country{China}}
\email{zhangwei.ltt@gmail.com}

\author{Xiaofeng He}
\affiliation{\institution{East China Normal University}
\country{China}}
\email{xfhe@cs.ecnu.edu.cn}

\author{Le Song}
\affiliation{\institution{Gatech}\country{USA}}
\email{lesong@cc.gatech.edu}

\author{Jingren Zhou}
\affiliation{\institution{Damo Academy, Alibaba Group}\country{China}}
\email{jingren.zhou@alibaba-inc.com}

\author{Hongxia Yang}
\affiliation{\institution{Damo Academy, Alibaba Group}\country{China}}
\email{yang.yhx@alibaba-inc.com}

\begin{abstract}

Dynamic graph modeling has recently attracted much attention due to its extensive applications in many real-world scenarios, such as recommendation systems, financial transactions, and social networks. Although many works have been proposed for dynamic graph modeling in recent years, effective and scalable models are yet to be developed.  In this paper, we propose a novel graph neural network approach, called TCL, which deals with the dynamically-evolving graph in a continuous-time fashion and enables effective dynamic node representation learning that captures both the temporal and topology information. Technically, our model contains three novel aspects. 
First, we generalize the vanilla Transformer to temporal graph learning scenarios and design a graph-topology-aware transformer. 
Secondly, on top of the proposed graph transformer, we introduce a two-stream encoder that separately extracts representations from temporal neighborhoods associated with the two interaction nodes and then utilizes a co-attentional transformer to model inter-dependencies at a semantic level. Lastly, we are inspired by the recently developed contrastive learning and propose to optimize our model by maximizing mutual information (MI) between the predictive representations of two future interaction nodes.
Benefiting from this, our dynamic representations can preserve high-level (or global) semantics about interactions and thus is robust to noisy interactions. To the best of our knowledge, this is the first attempt to apply contrastive learning to representation learning on dynamic graphs. We evaluate our model on four benchmark datasets for  interaction  prediction and experiment results demonstrate the superiority of our model.

\end{abstract}

\maketitle
\section{Introduction}

Representation learning on graphs is gaining increasing interests since it has exhibited great potentials in many real-world applications, ranging from e-commerce~\cite{zhao2019intentgc,he2020lightgcn}, drug discovery~\cite{you2018graph,zitnik2018modeling}, social networks~\cite{wu2018socialgcn,zhao2019intentgc} to financial transactions~\cite{khazane2019deeptrax}. Previous works on graph representation learning mainly focus on static settings where the topological structures are assumed fixed. However, graphs in practice are often constantly evolving, i.e., the nodes and their associated interactions (edges) can emerge and vanish over time. For instance, sequences of interactions such as following new friends, sharing news with friends on Twitter, or daily user-item purchasing interactions on Amazon can naturally yield a dynamic graph. The graph neural networks (GNNs) methods~\cite{defferrard2016convolutional,graphsage,graph_attention} tailored for static graphs usually perform poorly on such dynamic scenarios because of the inability of utilizing temporal evolutionary information that is lost in the static settings.

Dynamic graph modeling aims to learn a low-dimensional embedding for each node that can effectively encode temporal and structural properties on dynamic graphs. This is an appealing yet challenging task due to the sophisticated time-evolving graph structures. Several works have been proposed these days. According to the way that dynamic graphs are constructed, these works can be roughly divided into discrete-time methods and continuous-time methods. The former methods~\cite{pareja2020evolvegcn,goyal2018dyngem,sankar2020dysat} rely on discrete-time dynamic graph construction that approximates dynamic graph as a series of graph snapshots over time. Usually, static graph encoding techniques like GCN~\cite{graphsage} or GAT~\cite{graph_attention} are applied to each snapshot, and then a recurrent neural network (RNN)~\cite{hochreiter1997long} or a self-attention mechanism~\cite{vaswani2017attention} is introduced to capture complicated temporal dependency among snapshots. However, the discrete-time methods can be sub-optimal since they ignore the fine-grained temporal and structural information which might be critical in real-world applications. Meanwhile, it is unclear how to specify the size of the time intervals in different applications. To tackle these challenges, the latter continuous-time methods~\cite{rossi2020temporal,xu2020inductive,dai2016deepcoev,jodie,chang2020continuous} are considered and have achieved state-of-the-art results. Methods in this direction focus on designing different temporal neighborhood aggregation techniques applied to the fine-grained temporal graphs which are represented as a sequence of temporally ordered interactions. For instance, TGAT~\cite{xu2020inductive} proposes a continuous-time kernel encoder combined with a self-attention mechanism to aggregate information from temporal neighborhood. TGNs~\cite{rossi2020temporal} introduces a generic temporal aggregation framework with a node-wise memory mechanism. Chang et. al. propose a dynamic message passing neural network to capture the high-order graph proximity on their temporal dependency interaction graph (TDIG)~\cite{chang2020continuous}.

Although continuous-time methods have achieved impressive results, there exist some limitations: \textbf{First},
when aggregating information from temporal neighborhoods of the two target interaction nodes, most of the aforementioned-methods~\cite{jodie,chang2020continuous,dai2016deepcoev} employ RNNs-like architectures. Such methods suffer from vanishing gradient in optimization and are unable to capture long-term dependencies. The learned dynamic representations will degrade especially when applied to complicated temporal graphs. \textbf{Secondly}, these methods typically compute dynamic embeddings of the two target interactions nodes separately without considering the semantic relatedness between their temporal neighborhoods (i.e. history behaviors), which may also be a causal factor for the target interaction. For example, in a temporal co-author network, the fact is that nodes A and B previously co-authored with node C respectively can promote a new potential collaboration between nodes A and B. Therefore, modeling mutual influences between the two temporal neighborhoods can aid informative dynamic representations. \textbf{Lastly}, in optimization, most prior works typically model exact future by reconstructing future states~\cite{jodie} or leveraging a Temporal Point Process (TPP) framework~\cite{dai2016deepcoev,chang2020continuous} to model complicated stochastic processes of future interactions. However, they may learn the noisy information when trying to fit the next interactions. Besides, computing the survival function of an intensity function in TPP-based methods is expensive when the integration cannot be computed in closed-form.

To address the above limitations, we propose a novel continuous-time Transformer-based dynamic graph modeling framework via contrastive learning, called TCL. The main contributions of our work are summarized as follows:
\begin{itemize}[leftmargin=*]
\item We generalize the vanilla Transformer and enable it to handle temporal-topological information on the dynamic graph represented as a sequence of temporally cascaded interactions.
\item To obtain informative dynamic embeddings, We design a two-stream encoder that separately processes temporal neighborhoods associated with the two target interaction nodes by our graph-topology-aware Transformer and then integrate them at a semantic level through a co-attentional Transformer.
\item To ensure robust learning, we leverage a contrastive learning strategy that maximizes the mutual information (MI) between the predictive representations of future interaction nodes. To the our best knowledge, this is the first attempt to apply contrastive learning to dynamic graph modeling.

\item Our model is evaluated on four diverse interaction datasets for interaction prediction. Experimental results demonstrate that our method yields consistent and significant improvements over state-of-the-art baselines.
\end{itemize}
\section{Related Work}
This section reviews state-of-the-art approaches for dynamic graph learning. Since CL loss is used as our optimization objective, we also review recent works on CL-based graph representation learning.
\subsection{Dynamic Graph Modeling}
According to how the dynamic graph is constructed, we roughly divide the existing modeling approaches into two categories: discrete-time methods and continuous-time methods.

\noindent\textbf{Discrete-time Methods}. 
Methods in this category deal with a sequence of discretized graph snapshots that coarsely approximates a time-evolving graph.  The authors in~\cite{zhou2018dynamic} utilize temporally regularized weights to enforce the smoothness of nodes' dynamic embeddings from adjacent snapshots. However, this method may break down when nodes exhibit significantly varying evolutionary behaviors. DynGEM~\cite{goyal2018dyngem} is an autoencoding approach that minimizes the reconstruction loss and learns incremental node embeddings through initialization from the previous time steps. However, this method may not capture the long-term graph similarities. Inspired by the self-attention mechanism~\cite{vaswani2017attention}, DySAT~\cite{sankar2020dysat} computes dynamic embeddings by employing structural attention layers on each snapshot followed by temporal attention layers to capture temporal variations among snapshots. Recently, EvolveGCN~\cite{pareja2020evolvegcn} leverages RNNs to regulate the GCN model (i.e., network parameters) at every time step to capture the dynamism in the evolving network parameters. Despite progress, the snapshots-based methods inevitably fail to capture the fine-grained temporal and structural information due to the coarse approximation of continuous-time graphs. It is also challenging to specify a suitable aggregation granularity.

\noindent\textbf{Continuous-time Methods}. 
Methods in this category directly operate on time-evolving graphs without time discretization and focus on designing different temporal aggregators to extract information. The dynamic graphs are represented as a series of chronological interactions with precise timestamps recorded. DeepCoevolve~\cite{dai2016deepcoev} and its variant JODIE~\cite{jodie} employ two coupled RNNs to update dynamic node embeddings given each interaction. They provide an implicit way to construct the dynamic graph where only the historical interaction information of the two involved nodes of the interactions at time $t$ are utilized. The drawbacks are that they are limited to modeling first-order proximity while ignoring the higher-order temporal neighborhood structures. To exploit the topology structure of the temporal graph explicitly, TDIG-MPNN~\cite{chang2020continuous} proposes a graph construction method, named Temporal Dependency Interaction Graph (TDIG), which generalizes the above implicit construction and is constructed from a sequence of cascaded interactions. Based on the topology of TDIG, they employ a graph-informed Long Short Term Memory (LSTM)~\cite{hochreiter1997lstm} to obtain the dynamic embeddings. However, the downside of the above methods is that it is not good at capturing long-range dependencies and is difficult to train, which is also the intrinsic weaknesses of RNNs. Recent work TGAT~\cite{xu2020inductive} and TGNs~\cite{rossi2020temporal} adopt a different graph construction technique, i.e., a time-recorded multi-graph, which accommodates more than one interaction (edge) between a pair of nodes. TGAT uses a time encoding kernel combined with a graph attention layer~\cite{graph_attention} to aggregate temporal neighbors. Just like the encoding process in static models (e.g., GraphSAGE~\cite{graphsage}), a single TGAT layer is used to aggregate one-hop neighborhoods and by stacking several TGAT layers, it can capture high-order topological information. TGNs generalizes the aggregation of TGAT and utilizes a node-wise memory to capture long-term dependency.


\subsection{Contrastive Learning On Graph Representation Learning}  
Recently, state-of-the-art results in unsupervised graph representation learning have been achieved by leveraging a contrastive learning loss that contrasts samples from a distribution that contains dependencies of interest and the distribution that does not. Deep graph Infomax~\cite{velickovic2019deep} learns node representations by contrasting representations of nodes that belong to a graph and nodes coming from a corrupted graph. InfoGraph~\cite{sun2019infograph} learns graph-level representations by contrasting the representations at graph level and that of sub-structures at different scales. Motivated by recent advances in multi-view contrastive learning for visual representation learning, ~\cite{hassani2020contrastive} proposes a contrastive multi-view representation learning at both node and graph levels. According to~\cite{poole2019variational}, contrastive objectives used in these methods can be seen as maximizing the lower bounds of MI. A recent study~\cite{tschannen2019mutual} has shown that the success of these models is not only attributed to the properties of MI but is also influenced by the choice of the encoder and the MI estimators.

\section{Preliminaries}
To make our paper self-contained, we start with introducing the basic knowledge of Temporal Dependency Interaction Graph~\cite{chang2020continuous}, Transformer~\cite{vaswani2017attention} and Contrastive Learning~\cite{oord2018representation}, upon which we build our new method. 


\begin{figure}[htb]
    \centering
    \includegraphics[width=1.0\linewidth]{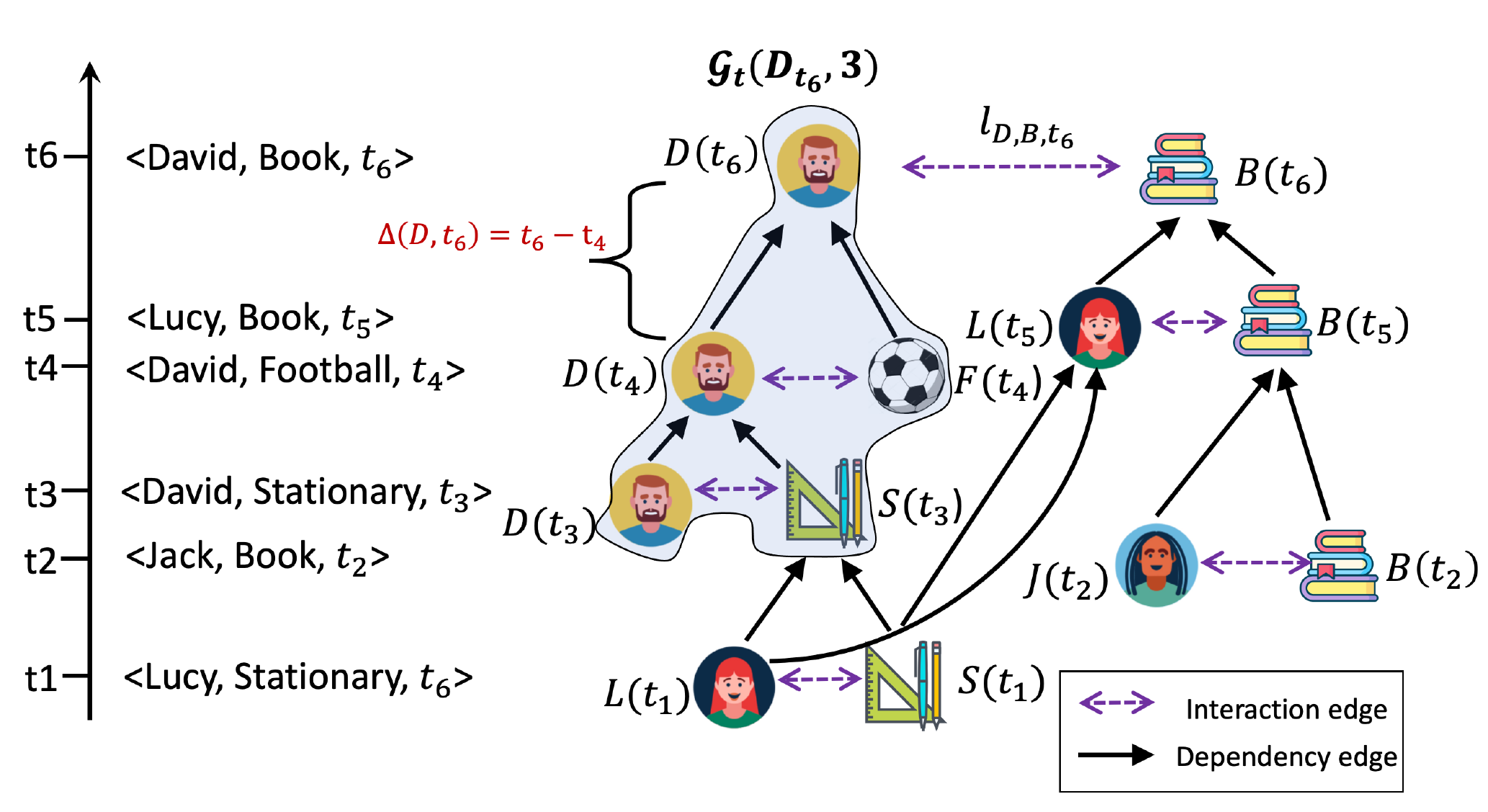}
    \vspace{-1em}
    \caption{Illustration of a Temporal Dependency Interaction Graph induced from a sequence of six chronological interactions.}
    \label{fig:tdig definition}
    \vspace{-0.1in}
\end{figure}
\subsection{Temporal Dependency Interaction Graph}
Temporal Dependency Interaction Graph (TDIG) proposed by ~\cite{chang2020continuous} is constructed by a sequence of temporally cascaded chronological interactions. Compared with other construction methods, TDIG as illustrated in Figure \ref{fig:tdig definition} maintains the fine-grained temporal and structural information of dynamic graphs. Therefore, in this paper, we select TDIG as our \textit{backbone modeling}.

Formally, a \emph{temporal dependency interaction graph} $\mathcal{G}_{t} = (\mathcal{V}_{t}, \mathcal{E}_{t})$ ($TDIG$) consists of a node set $\mathcal{V}_{t}$ and an edge set $\mathcal{E}_{t}$ indexed by time $t \in \mathbb{R}^{+}$ and is constructed based on a sequence of chronological interactions up to time $t$. An interaction occurring at time $t$ is denoted as $l_{u,v,t}$, where nodes $u$, $v \in \mathcal{V}_{t}$ are the two parties involved in this interaction. Since one node can have multiple interactions happening at different time points, for convenience, we let $u_t$ represent the node $u$ at time $t$ who was involved in $l_{u,v,t}$. There are two types of edges in $\mathcal{E}_{t}$. One is the \emph{interaction edge} that corresponds to an interaction $l_{u,v,t}$. The other is the \emph{dependency edge} that links the current node $u_t$ to the two dependency nodes that were involved in $u_t$'s last interaction at time $t_{u-}$ (just before time t). The \emph{dependency edge} represents the evolution and the causality between temporal relevant nodes. The two dependencies nodes of $u_t$ are denoted as $u(t_{u-})^{(1)}$ and $u(t_{u-})^{(2)}$ respectively. The time interval between $l_{u,v,t}$ and the node $u$'s last interaction, denoted as $\Delta_{(u,t)}=t-t_{u-}$, is treated as a dependency edge feature. To reduce the computational cost in practice, history information present in the sub-graphs rather than a whole TDIG is utilized to form dynamic embeddings. $\mathcal{G}_{t}(u,k)$ is denoted as the max $k$-depth sub-graph(i.e., temporal neighborhoods) rooted at $u_t$, where $k$ is a hyper-parameter. For more details, we refer the readers to the original paper~\cite{chang2020continuous} for a complete description.

\subsection{Transformer}
Transformer~\cite{vaswani2017attention} has achieved state-of-the-art performance and efficiency on many NLP tasks that have been previously dominated by
RNN/CNN-based~\cite{sutskever2014sequence,luong2015effective} approaches. 
A Transformer block relies heavily on a multi-head attention mechanism to learn the context-aware
representation for sequences, which can be defined as:
\begin{align}
\boldsymbol{O} &=\operatorname{MultiHead}(\boldsymbol{Q}, \boldsymbol{K}, \boldsymbol{V})=\operatorname{Concat}\left(\boldsymbol{O}_{1}, \boldsymbol{O}_{2}, \ldots, \boldsymbol{O}_{d_h}\right) \boldsymbol{W}_{O}
\end{align}
where $\boldsymbol{W}_{O}$ is a trainable parameter. $d_h$ is  the number of heads. $\boldsymbol{O}_{i}$ is computed as:
\begin{align}
  \boldsymbol{O}_{i}=\operatorname{Attention}\left( \boldsymbol{Q}\boldsymbol{W}_{Q, i}, \boldsymbol{K}\boldsymbol{W}_{K, i}, \boldsymbol{V}\boldsymbol{W}_{V, i} 
  \right)  
\end{align}

where $\boldsymbol{W}_{*,i} \in \mathbb{R}^{d \times d}$
are trainable parameters and $\operatorname{Attention}(\cdot,\cdot,\cdot)$ is the scaled dot-product attention defined as:
\begin{align}
\operatorname{Attention}(\boldsymbol{Q},\boldsymbol{K},\boldsymbol{V}) =\operatorname{softmax}\left(\frac{\boldsymbol{Q} \boldsymbol{K}^{{T}}}{\sqrt{d}}\right)\boldsymbol{V}
\end{align}
As input to Transformer, an element in a sequence is represented by an embedding vector. The multi-head attention mechanism works by injecting the Positional Embedding into the element embeddings to be aware of element orders in a sequence. 

\subsection{Contrastive Learning}
Contrastive learning approaches have attracted much attention for learning word embeddings in natural language models~\cite{klein2020contrastive}, image classification~\cite{chen2020simple}, and static graph modeling~\cite{velickovic2019deep}. Contrastive learning optimizes an objective that enforces the positive samples to stay in a close neighborhood and negative samples far apart:
\begin{align}
    \mathcal{L}_{\operatorname{contrast}}=-\mathbb{E}\left[ \log \frac{e^{\operatorname{Sim_{\theta}}\left(\mathbf{x}, \mathbf{x_{+}}\right)}}{e^{\operatorname{Sim_{\theta}}\left(\mathbf{x}, \mathbf{x_{+}}\right)}+\sum_{\mathbf{x_{-}}} e^{\operatorname{Sim_{\theta}}\left(\mathbf{x}, \mathbf{x_{-}}\right)}}\right]
\end{align}
where the positive pair $(\mathbf{x},\mathbf{x_{+}})$ is contrasted with negative pair $(\mathbf{x},\mathbf{x_{-}})$. A discriminating function $\operatorname{Sim}_{\theta}(\cdot)$ with parameters $\theta$ is trained to achieve a high value for congruent pairs and
low for incongruent pairs. According to the proofs~\cite{poole2019variational}, minimizing the objective $\mathcal{L}_{\operatorname{contrast}}$ can maximize the lower bounds of MI.

\section{Proposed Method}

Given the topology of TDIG, we first describe how to generalize the vanilla Transformer to handle temporal and structural information. To obtain informative dynamic embeddings, we design a two-stream encoder that first processes historical graphs associated with the two interaction nodes by a graph-topology-aware Transformer and then propose a co-attentional Transformer to fuse them at the level of semantics. Finally, we will introduce our temporal contrastive objective function used in model training. 
\subsection{Graph Transformer}
Considering the advantages of the Transformer in capturing long-term dependencies and in computational efficiency, we propose to extract temporal and structural information of TDIG by Transformer type of architecture. The vanilla Transformer architecture cannot be directly applied to dynamic graphs due to two reasons: (a) The vanilla Transformer is designed to process equally spaced sequences and cannot deal with the irregularly spaced time intervals between interactions that can be critical in analyzing the interactive behaviors~\cite{vaswani2017attention}. (b) The vanilla Transformer considers the attention with a flat structure, which is not suitable to capture the hierarchical and structured dependency relationships exhibited in TDIG. To address these challenges, we make two adaptions: (1) We leverage the depths of nodes together with the time intervals to encode nodes' positions or hierarchies. (2) We inject the graph structure into the attention mechanism by performing the masked operation. We will elaborate as follows.
\subsubsection{Embedding Layer} 
Let $\mathbf{S}_{u} =\left[s_{1}^{u}, \ldots,s_{i}^{u}, \ldots, s_{m}^{u}\right]$  be the set of nodes $\{s_{i}^{u}\}_{i=1, \dots, m}$ in $\mathcal{G}_{t}(u,k)$ where the nodes are listed according to a pre-determined order of graph traversal  (e.g., the breadth-first-search) and $m$ is the number of nodes in  $\mathcal{G}_{t}(u,k)$.  
We create a node embedding matrix $\mathbf{M} \in \mathbb{R}^{\left(|\mathcal{V}_{t}|\right) \times d}$, where each row of the matrix $\mathbf{M}$ corresponds to a d-dimensional embedding of a specific node in node set $\mathcal{V}_{t}$. We then retrieve the node embedding matrices for the node sequences $\mathbf{S}_{u}$:
\begin{align}
\mathbf{E}_{u}^{node}=\left[\begin{array}{c}
\mathbf{M}_{s_{1}^{u}} \\
\mathbf{M}_{s_{2}^{u}} \\
\cdots \\
\mathbf{M}_{s_{m}^{u}}
\end{array}\right]
\end{align}
where $\mathbf{E}_{u}^{node} \in \mathbb{R}^{ m \times d}$ is the concatenation
of the corresponding embeddings of nodes in $\mathbf{S}_{u}$.

\noindent \textbf{Positional Embedding:} Since the self-attention mechanism can not be aware of the nodes' positions or hierarchies on the TDIG, we first learn a nodes' depth embedding $\mathbf{P} \in \mathbb{R}^{ k \times d}$, where the $j$-th row of the matrix $\mathbf{P}$ denotes a d-dimensional embedding for the depth $j$. Denote
\begin{align}
\mathbf{E}_{u}^{depth}=\left[\begin{array}{c}
\mathbf{P}_{s_{1}^{u}} \\
\mathbf{P}_{s_{2}^{u}} \\
\cdots \\
\mathbf{P}_{s_{m}^{u}}
\end{array}\right]
\end{align}
where $\mathbf{E}_{u}^{depth} \in \mathbb{R}^{ m \times d}$ is the concatenation
of the corresponding depth embeddings of nodes in $\mathbf{S}_{u}$. The depths of nodes in the TDIG indicate temporal orders of the observed interactions in which these nodes are involved. However, only considering the depths of nodes is not enough, because the nodes in the TDIG have multiple instances with the same depths. To enhance nodes' positional information, we also employ time intervals (i.e., dependency edge features) that usually convey important behaviour information. Specifically, we use a time-interval projection matrix $\mathbf{W}_{\Delta}   
\in \mathbb{R}^{ d \times 1}$ to get the following time-interval embedding matrices:
\begin{align}
\mathbf{E}_{u}^{time}=\left[\begin{array}{c}
\mathbf{W}_{\Delta}\Delta_{(s_{1}^{u},t)} \\
\mathbf{W}_{\Delta}\Delta_{(s_{2}^{u},t)} \\
\cdots \\
\mathbf{W}_{\Delta}\Delta_{(s_{m}^{u},t)}
\end{array}\right]
\end{align}
where $\mathbf{E_{u}^{time}} \in \mathbb{R}^{ m \times d}$ is the concatenation
of the corresponding time-interval embeddings of nodes in $\mathbf{S}_{u}$. 

Finally, by injecting the TDIG-based positional embedding into the above node embeddings, we get the input embedding matrices $\hat{\mathbf{E}}_{u}\in \mathbb{R}^{ m \times d}$ for the next structure-aware attention layer:
\begin{align}
    \hat{\mathbf{E}}_{u}=\mathbf{E}_{u}^{node}+\mathbf{E}_{u}^{depth}+\mathbf{E}_{u}^{time} 
\end{align}

\subsubsection{Structure-Aware Attention Layer} 
The self-attention mechanism above used in the NLP domain usually allows every token to attend to every other token, but directly applying it to nodes of the TDIG will lead to a loss of the structural information. Instead, we propose to inject the TDIG structural information into the attention mask. Specifically, the structure attention mask $\boldsymbol{\Psi}^{u} \in \mathbb{R}^{ m \times m}$ for the subgraph $\mathcal{G}_{t}(u,k)$ is defined as:
\begin{align}
\boldsymbol{\Psi}^{u}_{i, j}=\left\{\begin{array}{ll}
0 & \text { if } k_{j} \in    \mathcal{SG}\left(q_{i}\right)   \\
-\infty & otherwise
\end{array}\right.
\end{align}
where $q_{i}$ and $k_{j}$ denote the query node indexed by $i$ and the key node indexed by $j$ respectively. The $\boldsymbol{\Psi}^{u}_{i, j}$ is set by 0 only for the affinity pairs whose key node $k_{j}$ belongs to the sub-graphs rooted at the query node $q_{i}$, i.e., $\mathcal{SG}\left(q_{i}\right)$. We then inject the structure attention mask $\boldsymbol{\Psi}^{u} \in \mathbb{R}^{ m \times m}$ into the scaled dot-product attention layer:
\begin{equation}
\begin{aligned}
&\boldsymbol{Q}_{u}^{b}, \boldsymbol{K}_{u}^{b}, \boldsymbol{V}_{u}^{b} =\boldsymbol{H}_{u}^{b} \boldsymbol{W}_{Q}, \boldsymbol{H}_{u}^{b} \boldsymbol{W}_{K}, \boldsymbol{H}_{u}^{b} \boldsymbol{W}_{V} \\
&\operatorname{M-Attention}(\boldsymbol{Q}_{u}^{b},\boldsymbol{K}_{u}^{b},\boldsymbol{V}_{u}^{b},\boldsymbol{\Psi}_{u}) =\operatorname{softmax}\left(\frac{\boldsymbol{Q}_{u}^{b} \boldsymbol({K}_{u}^{b})^{T}+\boldsymbol{\Psi}^{u}}{\sqrt{d}}\right)\boldsymbol{V}_{u}^{b}
\end{aligned}
\end{equation}
where $\boldsymbol{H}_{u}^{b}$ represents the $(b+1)$-th block input (when stacking B blocks) and $\boldsymbol{H}^{b=0}_{u}=\hat{\boldsymbol{E}_{u}}$. It could be observed that the attention output will be zero for the mask $\boldsymbol{\Psi}_{i,j}^{u}$ with negative infinity values (-$\infty$). In other words, each query node only attends to this query node' structural dependency nodes whose occurrence time are earlier than the occurrence time of this query node. The masked multi-head attention is denoted as 
\begin{align}
\operatorname{M-MultiHead}(\boldsymbol{Q}_{u}^{b},\boldsymbol{K}_{u}^{b},\boldsymbol{V}_{u}^{b},\boldsymbol{\Psi}^{u})
\end{align}
\subsubsection{Graph Transformer Block} 
Our graph-topology-aware Transformer block is defined with the following operation: \begin{align}
&\boldsymbol{O}_{u}^{b}=\operatorname{M-MultiHead}(\boldsymbol{Q}_{u}^{b-1},\boldsymbol{K}_{u}^{b-1},\boldsymbol{V}_{u}^{b-1},\boldsymbol{\Psi}^{u})
\\
&\boldsymbol{H}_{u}^{b}=\operatorname{LN}(\operatorname{FFN}(\operatorname{LN}(\boldsymbol{O}_{u}^{b}+\boldsymbol{Q}_{u}^{b}))+\operatorname{LN}(\boldsymbol{O}_{u}^{b}+\boldsymbol{Q}_{u}^{b}))
\end{align}
where "+" means a residual connection operation. $\operatorname{LN}$ denotes a layer normalization module~\cite{ba2016layer} which usually stabilizes the training process.
$\operatorname{FFN}$ represents a two-layer fully-connected feed-forward network module defined as:
\begin{align}
\operatorname{FFN}(x)=\sigma\left(x W_{1}+b_{1}\right) W_{2}+b_{2}
\end{align}
where $\sigma$ is an activation function usually implemented with a rectified linear unit~\cite{nair2010rectified}.
We sequentially apply these operations to get $\boldsymbol{H}_{u}^{b}$, which is the output of our graph Transformer block. 
\subsection{Encoders for Dynamic Embeddings }

\begin{figure}[tb]
    \centering
    \includegraphics[width=1.0\linewidth]{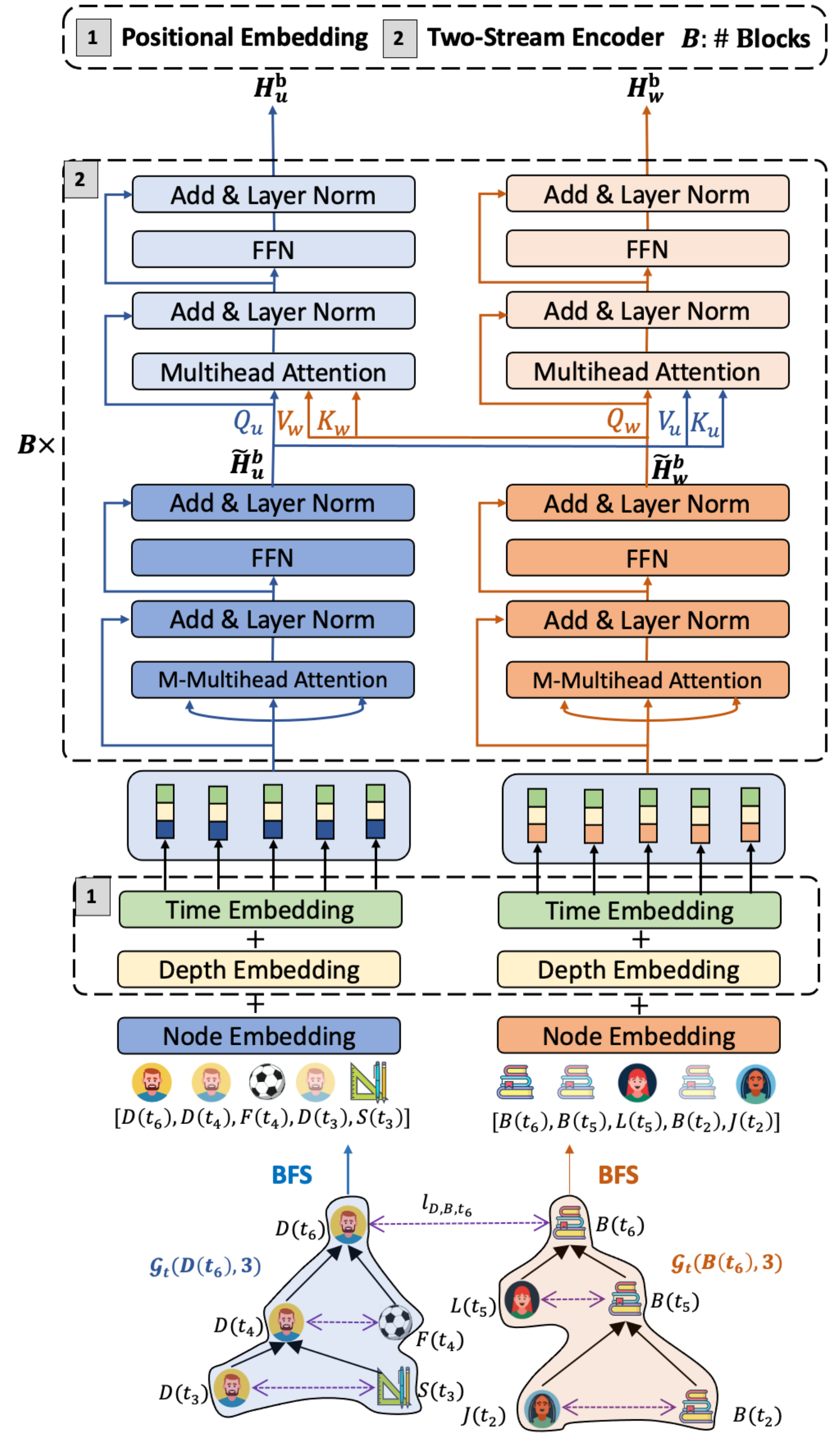}
    \vspace{-2em}
    \caption{Overview of Our Two-stream Encoder Framework.}
    \label{fig:tdig-encoder}
    \vspace{-0.3in}
\end{figure}

Given the topology of the $\mathcal{G}_{t}$, our work aims to obtain the dynamic embeddings of the interaction nodes $u(t)$ and $w(t)$, i.e., $\mathbf{h}_{u(t)} \in \mathbb{R}^{d}$ and $\mathbf{h}_{w(t)} \in \mathbb{R}^{d}$, by leveraging the history information in their sub-graphs $\mathcal{G}_{t}(u,k)$ and $\mathcal{G}_{t}(w,k)$. To achieve this, we devise an encoder with a two-stream architecture that first processes the $\mathcal{G}_{t}(u,k)$ and $\mathcal{G}_{t}(w,k)$ in two separate streams by the above graph-topology-aware Transformer layer and then interacts them at the level of semantics through a co-attentional Transformer layer. The encoder architecture is depicted in Fig.\ref{fig:tdig-encoder}
As shown in Fig. \ref{fig:tdig-encoder}, our two-stream encoder includes two successive parts. The first part is a graph Transformer block which we utilize to obtain intermediate embeddings of nodes in the $\mathcal{G}_{t}(u,k)$ and $\mathcal{G}_{t}(w,k)$, i.e., $\tilde{\boldsymbol{H}}^{b}_{u}$ and $\tilde{\boldsymbol{H}}^{b}_{w}$. The second part is a cross-attentional Transformer block that first models information correlation between $\tilde{\boldsymbol{H}}^{b}_{u}$ and $\tilde{\boldsymbol{H}}^{b}_{w}$, and then gets final informative dynamic embeddings matrices , ${\boldsymbol{H}}^{b}_{u}$ and ${\boldsymbol{H}}^{b}_{w}$, where we retrieval dynamic embeddings $\mathbf{h}_{u(t)}$ and $\mathbf{h}_{w(t)}$ for the interactions nodes $u(t)$ and $w(t)$.
The core of the cross-attentional Transformer block is a cross-attention operation by exchanging key-value pairs in the multi-headed attention. Formally, we have the following equations:
\begin{align}
&\boldsymbol{Q}_{u}^{b}, \boldsymbol{K}_{u}^{b}, \boldsymbol{V}_{u}^{b} =\tilde{\boldsymbol{H}}_{u}^{b} \boldsymbol{W}_{Q}, \tilde{\boldsymbol{H}}_{u}^{b} \boldsymbol{W}_{K}, \tilde{\boldsymbol{H}}_{u}^{b} \boldsymbol{W}_{V} \label{equ:co1} \\
&\boldsymbol{Q}_{w}^{b}, \boldsymbol{K}_{w}^{b}, \boldsymbol{V}_{w}^{b} =\tilde{\boldsymbol{H}}_{w}^{b}\boldsymbol{W}_{Q}, \tilde{\boldsymbol{H}}_{w}^{b}\boldsymbol{W}_{K}, \tilde{\boldsymbol{H}}_{w}^{b}\boldsymbol{W}_{V} \label{equ:co2} \\
&\boldsymbol{O}_{u}^{b}=\operatorname{MultiHead}(\boldsymbol{Q}_{u}^{b}, \boldsymbol{K}_{w}^{b}, \boldsymbol{V}_{w}^{b}) \\
&\boldsymbol{O}_{w}^{b}=\operatorname{MultiHead}(\boldsymbol{Q}_{w}^{b}, \boldsymbol{K}_{u}^{b}, \boldsymbol{V}_{u}^{b}) \\
&\boldsymbol{H}_{u}^{b}=\operatorname{LN}(\operatorname{FFN}(\operatorname{LN}(\boldsymbol{O}_{u}^{b}+\boldsymbol{Q}_{u}^{b}))+\operatorname{LN}(\boldsymbol{O}_{u}^{b}+\boldsymbol{Q}_{u}^{b}))\\
&\boldsymbol{H}_{w}^{b}=\operatorname{LN}(\operatorname{FFN}(\operatorname{LN}(\boldsymbol{O}_{w}^{b}+\boldsymbol{Q}_{w}^{b}))+\operatorname{LN}(\boldsymbol{O}_{w}^{b}+\boldsymbol{Q}_{w}^{b}))
\end{align}
The co-attention operations by exchanging key-value pairs in Eq.\ref{equ:co1} and Eq.\ref{equ:co2} enable our encoder to highlight relevant semantics shared by the two sub-graphs $\mathcal{G}_{t}(u,k)$ and $\mathcal{G}_{t}(w,k)$, and suppress the irrelevant semantics. Meanwhile, the attentive information from $\mathcal{G}_{t}(w,k)$ is incorporated into the final representation $\boldsymbol{H}^{b}_{u}$ and vice versa. By doing so, the obtained embeddings matrices $\boldsymbol{H}^{b}_{u}$ and $\boldsymbol{H}^{b}_{w}$ can be semantic and informative. We then retrieval dynamic node embeddings $\mathbf{h}_{u(t)}$ and $\mathbf{h}_{w(t)}$ for the two interactions nodes $u(t)$ and $w(t)$ from $\boldsymbol{H}^{b}_{u}$ and $\boldsymbol{H}^{b}_{w}$.

\noindent \textbf{Relation to previous work.} In comparison with the RNN-based information aggregation mechanism, such as JODIE, DeepCoevolve and TDIG-MPNN, our Transformer-based model excels in capturing the complex dependency structures over sophisticated time-evolving graphs. It successfully tackles the vanishing gradient issues inherited from RNN and eases optimization. What's more, all these previous methods learn node embeddings separately without considering the semantic relatedness between their temporal neighborhoods (i.e. history behaviors), which may also trigger new interactions. In contrast, we design a two-stream encoder to separately process temporal neighborhoods of the interaction nodes and integrate them at a semantic level through a co-attentional Transformer. As a result, we obtain more informative dynamic embeddings.

\subsection{Dynamic Graph Contrastive Learning}
For many generative time series models, the training strategies are formulated to maximize the prediction accuracy. 
For example, JODIE exploits the local smoothness of the data and optimizes the model by minimizing the $L_2$ distance between the predicted node embedding and the ground truth node’s embedding at every interaction. 
Based on the TPP framework, methods like DeepCoevolve~\cite{dai2016deepcoev} and TDIG-MPNN~\cite{chang2020continuous} model the occurrence rate of new interaction at any time $t$ and optimize the models by maximizing the likelihood.
However, their training performance relies on the modeling accuracy of future interactions and could be vulnerable to noise.  Regarding this, instead of reconstructing the exact future interactions or employing the complicated generative model to learn the stochastic processes of future interactions, we attempt to maximize the mutual information between the latent representations of interaction nodes in the future. In this way, we can learn the underlying latent representations that the interaction nodes have in common. 
Moreover, these underlying latent representations can preserve the high-level semantics of interactions and focus less on the low-level details, which are robust to noise information~\cite{wang2020understanding}. We apply the contrastive objective function to our dynamic embedding learning. According to proofs in ~\cite{poole2019variational}, minimizing this training objective function is equivalent to maximizing the lower bound of the mutual information between interaction nodes.

\subsubsection{Future Prediction}
Since we are leveraging the future interaction as our supervisory signal, we first construct the predictive future states of nodes $u$ and $v$ associated with future interaction $l_{u,v,t}$. Considering one node has two dependency nodes in TDIG, the predictive node' representation at future time $t$ is constructed based on the dynamic embeddings of this node' two dependency nodes involved in the previous interaction just before time $t$. Formally, we have the following equations:
\begin{align}
    &\bar{\mathbf{h}}_{{u}(t)}=\phi( [ \mathbf{h}_{{u^{(1)}(t_{u-})}}; \mathbf{h}_{{u^{(2)}(t_{u-})}}  ]) \\
    &\bar{\mathbf{h}}_{{v}(t)}=\phi( [\mathbf{h}_{{v^{(1)}(t_{v-})}}; \mathbf{h}_{{v^{(2)}(t_{v-})}}])
    \label{predictive function}
\end{align}
where $[;]$ denotes concatenation. $\phi(\cdot)$ is a projection head used to predict the future node representation. $\mathbf{h}_{{u^{(1)}(t_{u-})}}$ and $\mathbf{h}_{{u^{(2)}(t_{u-})}}$ are the dynamic embeddings of $u(t)$' two dependency nodes ${u^{(1)}(t_{u-})}$ and ${u^{(2)}(t_{u-})}$ respectively, which are obtained by extracting the history information from sub-graphs $\mathcal{G}_{t_{u-}}(u^{(1)},k)$ and $\mathcal{G}_{t_{u-}}(u^{(2)},k)$ using our two-stream encoder. 
The $\phi(\cdot)$ function takes the concatenation of $\mathbf{h}_{{u^{(1)}(t_{u-})}}$ and $\mathbf{h}_{{u^{(2)}(t_{u-})}}$ as input to forecast future representation $\bar{\mathbf{h}}_{{u}(t)} \in \mathbb{R}^{d}$.
Similarly, we have the predictive representation $\bar{\mathbf{h}}_{{v}(t)} \in  \mathbb{R}^{d}$ for the node $v$ at future time $t$.
In practice, $\phi(\cdot)$ is an $\operatorname{MLP}$ with two hidden layers and $\operatorname{PReLU}$ non-linearity.

\subsubsection{Contrastive Objective Function}
\begin{figure*}[htb]
    \centering
    \includegraphics[width=0.65\linewidth]{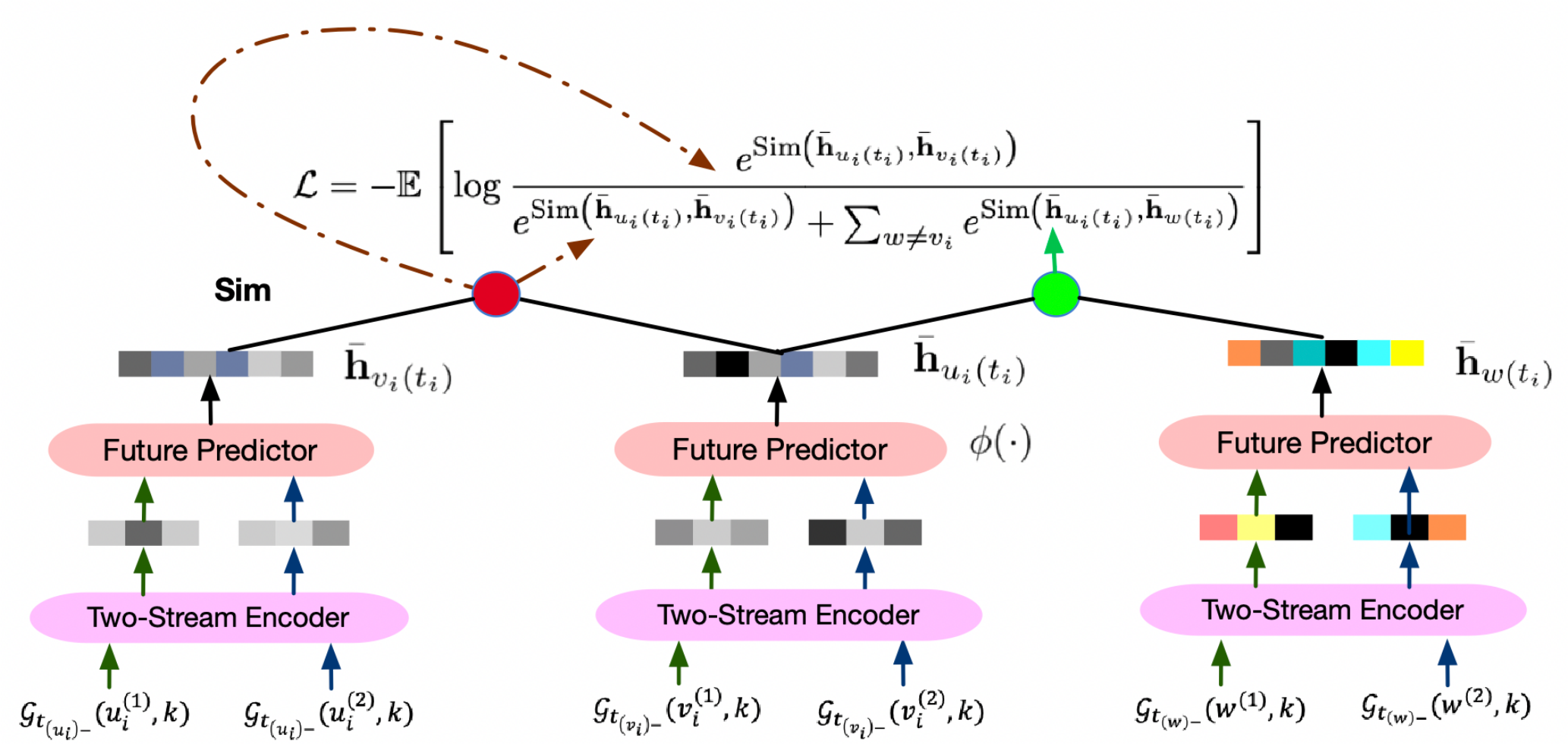}
    \vspace{-1em}
    \caption{Illustration of CL-based optimization strategy for proposed two-stream encoder.}
    \label{fig:procedure}
\end{figure*}

To train our two-stream encoders in an end-to-end fashion and learn the informative dynamic embeddings that aid future interaction prediction, we utilize the future interaction as our supervisory signal and maximize the mutual information between two nodes involved in the future interaction by contrasting their predictive representations. A schematic overview of our proposed method is in Fig.~\ref{fig:procedure}

Given a list of interactions $\mathcal{D}=\left\{\left(u_{i}, v_{i}, t_{i}\right)\right\}_{i=1}^{N}$ observed in a time window $[0, T]$, our contrastive learning is to minimize:
\begin{align}
    \mathcal{L}=-\mathbb{E}\left[ \log \frac{e^{\operatorname{Sim}\left(\bar{\mathbf{h}}_{u_{i}(t_{i})}, \bar{\mathbf{h}}_{v_{i}(t_{i})}\right)}}{e^{\operatorname{Sim}\left(\bar{\mathbf{h}}_{u_{i}(t_{i})}, \bar{\mathbf{h}}_{v_{i}(t_{i})}\right)}+\sum_{ w \neq v_{i}} e^{\operatorname{Sim}\left(\bar{\mathbf{h}}_{u_{i}(t_{i})}, \bar{\mathbf{h}}_{w(t_{i})}\right)}}\right]
    \label{clloss}
\end{align}
where $\operatorname{Sim(\cdot)}: \mathbb{R}^{d} \times \mathbb{R}^{d} \longmapsto \mathbb{R}$ denotes discriminator function that
takes two predictive representations as the input and then scores the agreement between them. Our discriminator function $\operatorname{Sim(\cdot)}$ is designed to explore both the additive and multiplicative interaction relations between $\bar{\mathbf{h}}_{u_{i}(t_{i})}$ and $\bar{\mathbf{h}}_{v_{i}(t_{i})}$:
\begin{align}
   \operatorname{Sim}\left(\bar{\mathbf{h}}_{u_{i}(t_i)}, \bar{\mathbf{h}}_{v_{i}(t_{i})}\right)=\text{SoftPlus} \left(  \mathbf{w}_{add}^{\top}(\bar{\mathbf{h}}_{u_{i}(t_{i})} +\bar{\mathbf{h}}_{v_{i}(t_{i})}) \right.  \nonumber\\
    \left.
    {+\mathbf{w}_{mul}^{\top}(\bar{ \mathbf{h}}_{u_{i}(t_{i})} \odot\bar{\mathbf{h}}_{v_{i}(t_{i})} }\right)
\end{align}
Essentially, the contrastive objective
function acts as a multi-way classifier that distinguishes the positive pairs
out of all other negative pairs. In our case, the positive pair is ($\bar{\mathbf{h}}_{u_{i}(t_{i})}, \bar{\mathbf{h}}_{v_{i}(t_{i})}$), i.e., the predictive representations of two nodes $u_{i}$ and $v_{i}$ involved in an interaction at time $t_{i}$. 
All the other pairs ($\bar{\mathbf{h}}_{u_{i}(t_{i})}, \bar{\mathbf{h}}_{w(t_{i})}$) where $w \neq v_{i}$ are negative pairs, which means all the other items that don't have interactions with $u_{i}$ at time $t_{i}$ will be treated as negative items. The aim of our contrastive loss here is to train our two-stream encoder that maximises the shared information between positive pairs, while minimizing the shared information between negative pairs that are well separated. Compared with existing training methods that model the exact future, our proposed contrastive loss makes the obtained dynamic embeddings more informative and robust to noise information. A schematic overview of our training method is shown in Fig.~\ref{fig:procedure}. We describe the pseudocode of training in Appendix. 

\section{Experiments}
We evaluate the effectiveness of TCL by comparing with baselines on four diverse interaction datasets. Meanwhile, the ablation study is conducted to understand which sub-module of TCL contributes most to the overall performance. We also conduct experiments of parameter sensitivity.
\subsection{Experimental Setting}

\subsubsection{Datasets}
\label{sec:exp}
We evaluate our
proposed model for temporal interaction prediction on four diverse interaction datasets as shown in Table~\ref{tab:data_stat}.

\noindent\textbf{CollegeMsg}~\cite{snapnets}. This dataset consists of message sending activities on a social network at the University of California, Irvine. The dataset we use here contains 999 senders and 1,352 receivers, with totally 48,771 activities. 

\noindent\textbf{Reddit}~\cite{jodie}. This public dataset consists of one month of posts made by users on subreddits. We select the $984$ most active subreddits as items and the $9,986$ most active users. This results in $672,447$ interactions.

\noindent\textbf{LastFM}~\cite{snapnets}. We use the dataset from top 1000 active users and 1000 popular songs, where the interactions indicate the user listen to the song.

\noindent\textbf{Wikipedia}~\cite{snapnets}. We use the data from 998 frequent edited pages and 7619 active users on Wikipedia, yielding totally 128271 interactions.

These four datasets are from different real-world scenes and vary in terms of the number of interactions, interaction repetition density and time duration, where interaction repetition density is the number of the intersection of training set and test set divides by the number of test set.

\begin{table}
\small
\centering
\setlength\abovecaptionskip{2pt}
\caption{Dataset Statistics.}
\setlength{\tabcolsep}{2pt}
\begin{tabular}{lcccc}  
\toprule
Dataset     & CollegeMsg & Wikipedia& LastFM& Reddit  \\
\midrule
\#u      & 999      & 7619  & 1000 & 9986       \\
\#v      & 1352    & 998 & 1000 & 984       \\
\#Interactions    & 48,771   & 128,271 & 1,293,103& 672,447     \\
Interaction Repetition Density      & 67.28\%   & 87.78\% & 88.01\% & 53.97\%   \\
Duration (day) & 193.63 & 30.00& 1586.89& 3882.00  \\
\bottomrule
\end{tabular}
\label{tab:data_stat}
\vspace{-1em}
\end{table}

\begin{table*}[ht]
\small
\centering
\setlength\abovecaptionskip{2pt}
\caption{Overall Comparison on Temporal Interaction Prediction.}
\setlength{\tabcolsep}{4pt}
\resizebox{\textwidth}{!}{
\begin{tabular}{c|c|cccccccc}  
\toprule
\multirow{2}{*}{\textbf{Group} } &\multirow{2}{*}{\textbf{Method} }    & \multicolumn{2}{c}{\textbf{CollegeMsg}} & \multicolumn{2}{c}{\textbf{Wikipedia}} & \multicolumn{2}{c}{\textbf{LastFM}} & \multicolumn{2}{c}{\textbf{Reddit}} \\ 
\cline{3-10}
& & MR$\downarrow$ & Hit$@$10$\uparrow$ & MR$\downarrow$ & Hit$@$10$\uparrow$ & 
MR$\downarrow$ & Hit$@$10$\uparrow$ 
& MR$\downarrow$ & Hit$@$10$\uparrow$\\\hline
\textbf{Static}&GraphSage*
\cite{graphsage,graph_attention} 
&269.35$\pm$6.12&11.56$\pm$1.27
& 165.39$\pm$3.59&33.47$\pm$3.81 
& 261.67$\pm$2.17&11.02$\pm$1.04
&50.49$\pm$ 1.14&58.63$\pm$1.54
\\\hline
\multirow{3}{*}{\textbf{Discrete-time}}&CTDNE\cite{ctdne2018} 
&607.27$\pm$10.65 &1.85$\pm$0.53 
&350.68$\pm$6.80&11.89$\pm$0.62
&461.59$\pm$7.49&1.17$\pm$0.05
&51.30$\pm$1.37&58.55$\pm$1.08
\\
&DynGEM\cite{goyal2018dyngem} 
&337.64$\pm$9.30& 3.47$\pm$0.38
& 216.56$\pm$5.27&16.79$\pm$0.21
& 457.68$\pm$8.37&3.67$\pm$0.04
&52.43$\pm$1.75&45.37$\pm$0.90
\\
&DySAT\cite{sankar2020dysat}
&354.35$\pm$8.47& 2.69$\pm$0.47 
& 146.59$\pm$2.36 &17.64$\pm$0.89
& 292.91$\pm$4.28&4.63$\pm$0.36
&55.79$\pm$1.78&56.150$\pm$1.14

\\
\hline

\multirow{6}{*}{
\makecell[c]{\bf{Continuous-}\\\bf{time}}}&JODIE\cite{jodie}  
&286.62$\pm$7.14&24.09$\pm$0.92
&125.88$\pm$1.96&41.27$\pm$1.03
&289.53$\pm$4.91&27.97 $\pm$1.35 
&56.13$\pm$1.78&59.11$\pm$1.94
\\
&DeepCoevolve
\cite{dai2016deepcoev}
&439.63$\pm$11.71&2.53$\pm$0.21
&227.74$\pm$6.41&15.37$\pm$0.26
&376.23$\pm$5.38&20.44$\pm$0.68
&49.70$\pm$1.53&62.06$\pm$1.02
\\
&TGAT\cite{xu2020inductive}
&206.79$\pm$6.37&35.49$\pm$1.15
&87.03$\pm$2.41&72.05$\pm$1.37 
&199.24$\pm$2.51&30.27$\pm$0.92 
&\underline{45.11}$\pm$2.19&64.33$\pm$2.66
\\
&TGNs\cite{rossi2020temporal} 
&196.69$\pm$4.36&36.07$\pm$1.39 
&75.47$\pm$2.02&79.74$\pm$1.67 
&192.36$\pm$3.02&31.07$\pm$0.85 
&56.34$\pm$2.04&61.61$\pm$2.73\\

&TDIG-MPNN~\cite{chang2020continuous} 
&\underline{160.89}$\pm$1.58&\underline{39.37}$\pm$1.31
&\underline{69.28}$\pm$1.34&\underline{79.82}$\pm$0.49
&\underline{180.07}$\pm$2.43&\underline{33.29}$\pm$1.63&{46.91}$\pm$0.49&\underline{71.46}$\pm$0.08
\\
&TCL
&\textbf{137.57}$\pm$1.70&\textbf{45.53}$\pm$0.37
&\textbf{54.33}$\pm$1.05&\textbf{81.17}$\pm$0.16
&\textbf{149.31}$\pm$0.72& \textbf{35.32}$\pm$0.48
&\textbf{33.95}$\pm$1.27&\textbf{75.95}$\pm$0.47
\\\hline
&Improvement&14.49\%&15.64\%&21.58\%&1.69\%&16.01\%&9.38\%&24.74\%&6.28\% \\
\bottomrule
\end{tabular}}
\label{tab:Overall Performance}
\end{table*}




\subsubsection{Baselines}
We compare TCL with the following algorithms spanning three algorithmic categories:

\begin{itemize}[leftmargin=*]
    \item \textbf{Static Methods}. We consider \textbf{GraphSAGE}~\cite{graphsage} with four aggregators, namely, \textbf{GCN}, \textbf{MEAN}, \textbf{MAX} and \textbf{LSTM}. A \textbf{GAT}~\cite{graph_attention} aggregator is also implemented base on the \textbf{GraphSAGE} framework~\cite{graphsage}. For convenience, we report the best results among the above five aggregators denoted as \textbf{GraphSage*}.
    \item \textbf{Discrete-Time Methods}. Three snapshot-based dynamic methods are considered here. \textbf{CTDNE} \cite{ctdne2018} is an extension to \textbf{DeepWalk}~\cite{perozzi2014deepwalk} with time constraint for sampling orders; \textbf{DynGEM}~\cite{goyal2018dyngem} is an autoencoding approach that minimizes the reconstruction loss and learns incremental node embeddings through initialization from the previous time steps; \textbf{DySAT}~\cite{sankar2020dysat} computes dynamic embeddings by using structural and temporal attentions.
    
   \item \textbf{Continuous-Time Methods}. Five continuous-time baselines are considered here. \textbf{DeepCoevolve}~\cite{dai2016deepcoev} learns the dynamic embedding via utilizing a TPP framework to estimate the intensity of two nodes interacting in the future time; \textbf{JODIE}\cite{jodie} utilizes two RNN models and a projection function to learn dynamic embeddings; \textbf{TDIG-MPNN}~\cite{chang2020continuous} proposes a dynamic message passing network combined a selection mechanism to obtain dynamic embeddings; \textbf{TGAT}\cite{xu2020inductive} proposes temporal graph attention layer to aggregate temporal-topological neighborhood features; \textbf{TGNs}\cite{rossi2020temporal} extends \textbf{TGAT} by adding a memory modules to capture long-term interaction relations. 
       
\end{itemize}

\subsubsection{Evaluation Metrics}
Given an interaction $l_{u,v,t}$, each method outputs the node $u$'s preference scores over all the items at time $t$ in test set. We sort scores in a descending order and record the rank of the paired node $v$. We report the average ranks for all interactions in test data, which is denoted as \textbf{Mean Rank}(MR). We also report the \textbf{Hit@10} defined as the proportion of times that a test tuple appears in the top 10.

\subsubsection{Reproducibility}
For \textbf{GraphSage}, \textbf{CTDNE}, \textbf{DynGEM}, \textbf{DySAT}, \textbf{JODIE}, \textbf{TGAT}, and \textbf{TGNs}, we use their open-source implementations. For \textbf{DeepCoevole} and \textbf{TDIG-MPNN}, we use the source codes provided by their authors.
More details about parameter settings and implementation can be found in the Appendix.


\subsection{Overall Performances}
In this section, we evaluate our method and the baselines on temporal interaction prediction task. All experiments are repeated five times and averaged results are reported in Table ~\ref{tab:Overall Performance}, from which we have the following observations: (1)TCL consistently outperforms all the competitors on all the datasets. The improvement of TCL over the second-best results(underlined in Table ~\ref{tab:Overall Performance}) are 14.49\%, 21.58\%, 16.01\% and 24.74\% respectively in terms of MR scores. The strong performance verifies the superiority of TCL. (2)On average, the continuous-time methods perform better than the static and discrete-time methods, which can be explained by the fact that the fine-grained temporal and structural information is critical for dynamic scenarios. (3)In some scenarios, the performance of discrete-time dynamic methods is not better than that of static methods as expected. Similar phenomenons have also been observed by previous work DyRep~\cite{trivedi2019dyrep} and TDIG-MPNN. A possible explanation is that it’s non-trivial to specify the appropriate aggregation granularity(i.e., the number of the snapshots) for these scenarios. (5)In some scenarios, the static methods GraphSage* perform competitive with the continuous-time baselines. One possible reason could be that there are many repetitive interactions in our datasets and recurring interaction information can help models predict easily, especially for static methods which can make full use of structural information. (6)TCL and recent work including TGAT, TGNs and TDIG-MPNN surpass JODIE and DeepCoevolve by a large margin, which indicates the importance of exploiting information from high-order temporal neighborhood(i.g., the k-depth sub-graph information used in TCL). (7) TCL performs relatively better than the recent work TGAT, TGNs and TDIG-MPNN. The reasons could be two folds. First, all these baseline methods treat temporal information aggregations of two interactions nodes separately without considering the semantic relatedness between their temporal neighborhoods (i.e. history behaviors), which may be a causal factor for the target interaction, while the proposed two-stream encoder can utilize the co-attentional Transformer to capture inter-dependencies at semantic level. Second, we use the contrastive learning loss as our optimization objective that enable our dynamic embeddings to preserve  high-level (or global)semantics about interactions which is robust to noise information.

\subsection{Ablation Study}
\begin{table}[t]
\footnotesize
\centering
\setlength\abovecaptionskip{3pt}
\caption{The comparison of TCL with its variants.}
\setlength{\tabcolsep}{5pt}
\resizebox{0.47\textwidth}{14mm}{
\begin{tabular}{cccccccc}  
\toprule
\textbf{Dataset}  &\textbf{CollegeMsg}   &\textbf{Wikipedia}& \textbf{LastFM}& \textbf{Reddit}  \\ 
\midrule

Metrics & MR$\downarrow$  & MR$\downarrow$&MR$\downarrow$&MR$\downarrow$  \\\hline
TCL w/o TE &140.20&54.49&157.86&34.09\\ 
TCL w/o DE &139.11&55.85&151.48&35.05\\
TCL w/o CA-Transformer&146.18&57.42&152.31& 35.39\\
Two-Stream-Encoder+TPP&143.31&62.78&153.28&40.61\\
\hline
TCL&136.13&52.97&149.27&33.67 \\
\bottomrule
\end{tabular}}
\label{tab:ablation}
\end{table}

We perform ablation studies on our TCL framework by removing specific module one at a time to explore their relative importance. The components validated in this section are the positional embedding, the cross-attentional Transformer, and the contrastive learning strategy.

\noindent\textbf{Positional Embedding}. 
We design the graph positional embedding with the aim of enhancing the positional information of nodes. To this end, the positional embedding module encodes both the time interval information and depth information. To evaluate their importance, we test the removal of time embedding (i.e., \textbf{TCL w/o TE}) and the removal of depth embedding (i.e., \textbf{TCL w/o DE}).
From the results shown in Table~\ref{tab:ablation}, it can be observed that the performance degrades when removing either TE or DE, confirming the effectiveness of both time intervals between adjacent interactions and depth of the nodes.

 \noindent\textbf{Cross-attentional Transformer}. 
To aid informative dynamic representations, we propose a two-stream encoder which includes a cross-attentional Transformer module to aggregate information from temporal neighborhood with the mutual influence captured. To verify the effectiveness, we test the removal of the cross-attentional Transformer (i.e., \textbf{TCL w/o CA-Transformer}) in our two-stream encoder. As shown in Table~\ref{tab:ablation}, we find that TCL with the default setting outperforms TCL w/o CA-Transformer over all datasets by 5.04\% on average, demonstrating the important role of the cross-attentional in our encoder.


\begin{figure*}[htb]
    \centering
    \includegraphics[width=0.8\linewidth]{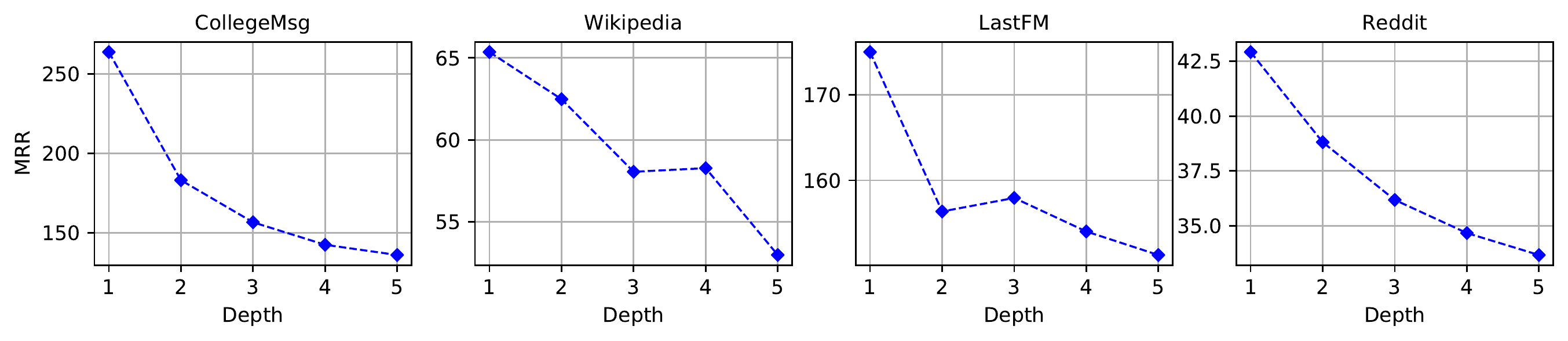}
    \setlength{\abovecaptionskip}{-1.5pt}%
    \caption{Performance of TCL w.r.t. different depths of the $k$-depth sub-graph.}
    \label{depthab}
\end{figure*}

\begin{figure*}[htb]
    \centering
    \includegraphics[width=0.8\linewidth]{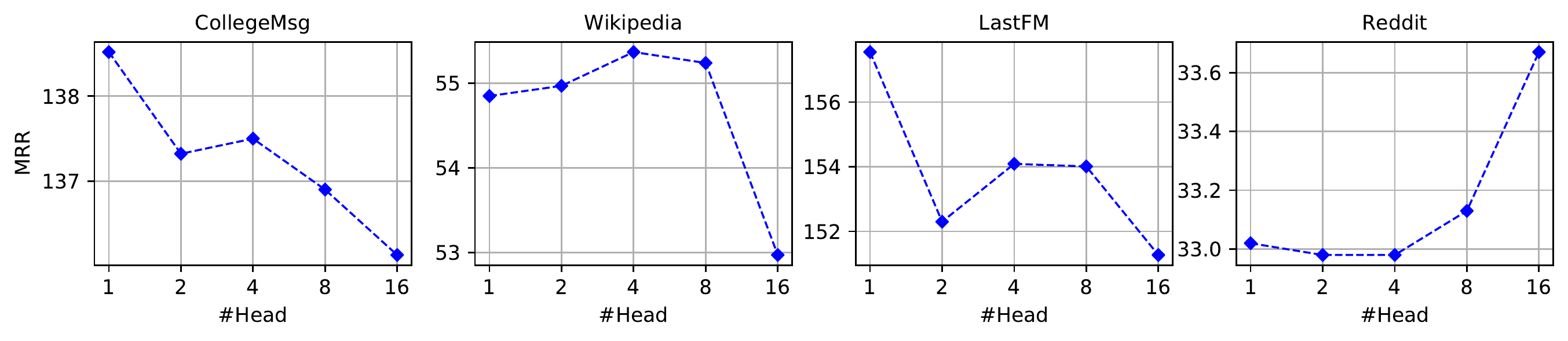}
     \setlength{\abovecaptionskip}{-1.5pt}%
    \caption{Performance of TCL w.r.t different number of heads.}
\vspace{-1em}
    \label{headab}
\end{figure*}

\noindent\textbf{Contrastive learning}. 
To improve robustness to noisy interactions, we utilize the contrastive learning as the objective function that maximizes the mutual information between the predictive representations of future interaction nodes.
To evaluate its effectiveness, we compare TCL with a variant that replaces the contrastive learning by a TPP objective (i.e., \textbf{Two-Stream-Encoder+TPP}). It can be observed that TCL outperforms Two-Stream-Encoder+TPP by a large margin, i.e., 9.76\% on average, which demonstrates the effectiveness of our optimization strategy.
\subsection{Parameter Sensitivity}
We investigate the impact of parameters on the future interaction prediction performance. 
\subsubsection{Impact of Sub-graph Depth}
We explore how the depth of the sub-graph impacts the performance. We plot the MR metric of TCL in different depth on four datasets. The results are summarized in Fig~\ref{depthab}. We find that the performance gradually improves with the increasing of depth, which verifies that exploiting information of high-order temporal neighborhood can benefit the performance.
\subsubsection{Impact of Attention Head Number}
Multi-head attention allows the model to jointly attend to information from different representation subspaces. We attempt to see how the attention head number in our two-stream encoder impacts the performance. We plot the MR metric with different number of heads in Fig~\ref{headab}. We observe that in most cases the performance improves when the head number increases, which demonstrates the effectiveness of multi-head attention. However, at some cases, more heads lead to degraded performance due to the possible over-fitting problem.

\section{Conclusion}

In this paper, we propose a novel continuous-time dynamic graph representation learning method, called TCL. TCL generalizes the vanilla Transformer and obtains temporal-topological information on dynamic graphs via a two-stream encoder.
The proposed contrastive learning can preserve the high-level semantics of interactions and focus less on the low-level details, which is robust to noise.
Extensive experiments verify the effectiveness and stability of TCL.
In the future, there are still two important problems to be considered, i.e., how to effectively model long-term history information on a time-evolving graph and how to scale well.

\bibliographystyle{ACM-Reference-Format}
\bibliography{sample-sigconf}
\clearpage
\appendix

\section{Reproducibility Supplement}


\subsection{Settings for Baselines}
To enhance the reproducibility of this paper, we first give the pseudocode of our training process. Then we describe the implementation of the baselines. Finally, we introduce the experimental environment and hyperparameters settings. 

\textbf{Implementation of Baselines}. 
We compare TCL with three categories of methods:
(1) Static Methods: GraphSage\footnote{https://github.com/williamleif/GraphSAGE}. 
(2) Discrete-Time Methods: CTDNE\footnote{https://github.com/stellargraph/stellargraph}, DynGEM\footnote{https://github.com/palash1992/DynamicGEM} and DySAT\footnote{https://github.com/aravindsankar28/DySAT}.
(3) Continuous-time Methods: JODIE\footnote{https://github.com/srijankr/jodie}, DeepCoevole, TGAT\footnote{https://github.com/StatsDLMathsRecomSys/Inductive-representation-learning-on-temporal-graphs}, TGNs\footnote{https://github.com/twitter-research/tgn} and TDIG-MPNN. 

For GraphSage*, the maximum number of 1/2/3/4/5-hop neighbor nodes is set to be 25/10/10/10/10. For discrete-time methods, we search the number of snapshots in \{1,5,10,15\} for all datasets. We search learning rates in \{0.0001, 0.0005, 0.001, 0.005, 0.01, 0.1\}, batchsize in \{128, 256, 512\} and keep the other hyper-parameters the same as their published version to obtain the best results of these methods. For continuous-time methods, we search the learning rates in \{0.0001, 0.0005, 0.001, 0.005\}, batch-size in \{128, 256, 512\} to obtain the best results of these methods and keep the other hyper-parameters same with their published version. We set the embedding dimension as $d=64$ for all the methods, maximum training epochs $20$ and $5$ negative samples (except for DynGEM and JODIE which do not require negative samples) for a fair comparison. 

\subsection{Settings for TCL }

\begin{table}[!h]
\small
\centering
\setlength\abovecaptionskip{2pt}
\caption{Hyper-parameters Settings.}
\setlength{\tabcolsep}{2pt}
\begin{tabular}{cc}  
\toprule
Hyper-parameters& Setting\\\hline
Learning rate & 0.0005\\
Optimizer & Adam\\
Mini-batch size & 512\\
Node embedding dimension & 64\\
Number of attention heads&16\\
The depth of the k sub-graph & 5\\
Dropout ratio in the input&0.6\\
Number Training Epoch&20\\
Number of negative samples&5\\ 
Number of blocks $B$&1\\
\bottomrule
\end{tabular}
\end{table}

\textbf{DataSet Split.}
We split CollegeMsg into training/validation/test sets by 60/20/20. For a fair comparison with JODIE, we split dataset LastFM by 80/10/10. For a fair comparison with TGNT and TGNs, we split Reddit and Wiki by 70/15/15.



\subsection{Pseudocode for TCL}
The pseudocode of the training procedure for TCL is detailed in Algorithm 1. 
\begin{algorithm}[h] 
\caption{TCL}
\begin{algorithmic}[1]
\REQUIRE The dynamic interaction set $\mathcal{D}=\{l_{u_i,v_i,t_i}\}_{i=1}^N$; Depth $k$; \# Heads $d_h$; \# Blocks $B$; \# Epochs $E$, number of Negative samples \# NS . Initialize the parameters of Two-Stream Encoder $Encoder(\cdot)$, and projection function $\phi(\cdot)$.
\FOR{$e$ in ${1,2,...,E}$} 
\FOR{($u_i,v_i,t_i$) in $\mathcal{D}$} 
\STATE Select \# NS negative items $w\neq v_{i}$ and extract the dependency k-depth sub-graphs of $u_i$, $v_i$, $w$ as \\ ($\mathcal{G}_{t_{u_i-}}(u_i^{(1)},k)$, $\mathcal{G}_{t_{u_i-}}(u_i^{(2)},k)$),\\ ($\mathcal{G}_{t_{v_i-}}(v_i^{(1)},k)$, $\mathcal{G}_{t_{v_i-}}(v_i^{(2)},k)$),\\ ($\mathcal{G}_{t_{w-}}(w^{(1)},k)$, $\mathcal{G}_{t_{w-}}(w^{(2)},k)$). 
\STATE Obtain the behavior sequences of each sub-graph via BFS,\\
$\mathbf{S}^{u_i^{(1)}} \leftarrow\mathcal{G}_{t_{u_i-}}(u_i^{(1)},k)$, 
$\mathbf{S}^{u_i^{(2)}} \leftarrow\mathcal{G}_{t_{u_i-}}(u_i^{(2)},k)$,\\
$\mathbf{S}^{v_i^{(1)}} \leftarrow\mathcal{G}_{t_{u_i-}}(v_i^{(1)},k)$,
$\mathbf{S}^{v_i^{(2)}} \leftarrow\mathcal{G}_{t_{u_i-}}(v_i^{(2)},k)$,\\
$\mathbf{S}^{w^{(1)}} \leftarrow\mathcal{G}_{t_{u_i-}}(w^{(1)},k)$,
$\mathbf{S}^{w^{(2)}} \leftarrow\mathcal{G}_{t_{u_i-}}(w^{(2)},k)$.
\STATE Encode each pair of dependency sub-graph via Two-Stream Encoder $Encoder$,\\
$\mathbf{h}_{{u_i^{(1)}(t_{u_i-})}}$,$\mathbf{h}_{{u_i^{(2)}}(t_{u_i-})} \leftarrow Encoder(\mathbf{S}^{u_i^{(1)}}, \mathbf{S}^{u_i^{(2)}}$),\\
$\mathbf{h}_{{v_i^{(1)}({t_{v_i-}})}}$,$ \mathbf{h}_{{v_i^{(2)}({t_{v_i-}})}} \leftarrow Encoder(\mathbf{S}^{v_i^{(1)}}, \mathbf{S}^{v_i^{(2)}}$),\\
$\mathbf{h}_{{w^{(1)}({t_{w-}})}}$,$ \mathbf{h}_{{w^{(2)}({t_{w-}})}} \leftarrow Encoder(\mathbf{S}^{w^{(1)}},\mathbf{S}^{w^{(2)}}$)
\STATE Project dependency node embeddings to node embedding via Future Prediction Function $\phi$,
\STATE
$\bar{\bf{h}}_{u_i(t)} \leftarrow \phi ([\bf{h}_{u_i^{(1)}(t_{u_{i-}})};\bf{h}_{u_i^{(2)}(t_{u_i-})}])$\\
$\bar{\bf{h}}_{v(t)} \leftarrow \phi([\bf{h}_{v_i^{(1)}(t_{v_i-})}; \bf{h}_{v_i^{(2)}(t_{v_i-})}$\\
$\bar{\mathbf{h}}_{{w}(t)} \leftarrow \phi([\mathbf{h}_{{w^{(1)}}(t_{w-})}; \mathbf{h}_{w^{(2)}(t_{w-})}])$
\STATE Calculate the CL loss according to Eq.~\ref{clloss} and update the parameters of $Encoder$ and $\phi$.
\ENDFOR
\ENDFOR
\end{algorithmic}
\end{algorithm}



\end{document}